\pdfoutput=1

\documentclass[11pt]{article}

\usepackage[whole]{bxcjkjatype} 
\usepackage[unicode]{hyperref}

\usepackage{acl}

\usepackage{times}
\usepackage{latexsym}

\usepackage[T1]{fontenc}

\usepackage{microtype}
\usepackage{inconsolata}
\usepackage{enumitem}

\usepackage{graphicx}

\usepackage{array}
\usepackage{here}
\usepackage{float}
\usepackage{multicol}

\usepackage{multirow,tabularx}
\usepackage{array}
\usepackage{here}
\usepackage{float}
\usepackage{multicol}
\usepackage{arydshln}

\usepackage{amsmath}
\usepackage{amssymb}
\usepackage{amsfonts}
\usepackage{latexsym}
\usepackage{comment}
\usepackage{color}
\usepackage{seqsplit}

\newcommand{\tabH}{\rule{0pt}{2.2ex}}
\newcommand{\bhline}{\noalign{\hrule height 1.2pt}}
\newcommand{\footlink}[1]{\footnote{\url{#1}}}

\newcommand{\foothreftext}[2]{\footnotetext{\href{#1}{\texttt{\seqsplit{#2}}}}}

\newcommand{\footsep}{{$^{,}$}}

\newcommand{\hf}[1]{\href{https://huggingface.co/#1}{#1}}

\newcommand{\RuriSmall}{{$\text{Ruri}_\text{small}$}}
\newcommand{\RuriBase}{{$\text{Ruri}_\text{base}$}}
\newcommand{\RuriLarge}{{$\text{Ruri}_\text{large}$}}

\newcommand{\RuriPTSmall}{{$\text{Ruri-PT}_\text{small}$}}
\newcommand{\RuriPTBase}{{$\text{Ruri-PT}_\text{base}$}}
\newcommand{\RuriPTLarge}{{$\text{Ruri-PT}_\text{large}$}}

\newcommand{\RuriRerankerSmall}{{$\text{Ruri-Reranker}_\text{small}$}}
\newcommand{\RuriRerankerBase}{{$\text{Ruri-Reranker}_\text{base}$}}
\newcommand{\RuriRerankerLarge}{{$\text{Ruri-Reranker}_\text{large}$}}

\usepackage{hyperref}

\title{Ruri: Japanese General Text Embeddings}

\author{
  Hayato Tsukagoshi \hspace{2em} Ryohei Sasano\\
  Graduate School of Informatics, Nagoya University\\
  \texttt{tsukagoshi.hayato.r2@s.mail.nagoya-u.ac.jp}, \\
  \texttt{sasano@i.nagoya-u.ac.jp} \\
}

\begin{document}
\maketitle
\begin{abstract}
We report the development of Ruri, a series of Japanese general text embedding models.
While the development of general-purpose text embedding models in English and multilingual contexts has been active in recent years, model development in Japanese remains insufficient.
The primary reasons for this are the lack of datasets and the absence of necessary expertise.
In this report, we provide a detailed account of the development process of Ruri.
Specifically, we discuss the training of embedding models using synthesized datasets generated by LLMs, the construction of the reranker for dataset filtering and knowledge distillation, and the performance evaluation of the resulting general-purpose text embedding models.
\end{abstract}

\renewcommand*{\arraystretch}{1.1}

\section{Introduction}

Text embeddings are widely used for tasks such as retrieval-augmented generation (RAG) and similar document retrieval~\cite{SentenceBERT,SimCSE,E5}. 
In recent years, the development of general-purpose text embedding models trained on diverse datasets has become increasingly common~\cite{E5,mE5,GTE,BGE,JinaBERT}. 
However, these efforts have mainly focused on English and multilingual models, where the proportion of Japanese vocabulary and training datasets is relatively small.
Building embedding models using large-scale Japanese datasets may enable the creation of higher-performing models.

In this report, we present a general-purpose text embedding model specialized for Japanese, which was developed through contrastive pre-training, the construction of synthetic training datasets using LLMs, and fine-tuning on high-quality datasets. 
Our contributions are summarized as follows:
\begin{enumerate}[leftmargin=0.5cm, itemsep=0.3pt]
    \item We collected datasets for building Japanese embedding models and released them under a permissive license.
    \item To address the lack of Japanese retrieval datasets, we constructed a synthetic dataset using LLMs. A performance comparison with and without the synthetic dataset in benchmark tests showed a difference of over 1 point, confirming its utility in training Japanese embedding models.
    \item We created a large-scale dataset for contrastive pre-training in Japanese, demonstrating its effectiveness by outperforming existing multilingual models, even when using contrastive pre-training alone.
    \item We developed a Japanese reranker, achieving the highest performance among existing Japanese rerankers.
    \item We built the Japanese embedding model Ruri, which significantly outperformed existing models in text embedding benchmarks.
\end{enumerate}
Our models and datasets are publicly available\footnotemark\foothreftext{https://huggingface.co/collections/cl-nagoya/ruri-japanese-general-text-embeddings-66cf1f3ee0c8028b89d85b5e}{https://huggingface.co/collections/cl-nagoya/ruri-japanese-general-text-embeddings-66cf1f3ee0c8028b89d85b5e}.

\section{Contrastive Pre-training}
\label{sec:contrastive-pre-training}

Recent research on text embeddings has seen a growing interest in a two-stage learning approach~\cite{E5,GTE,BGE}.
This approach consists of two steps: contrastive pre-training and fine-tuning.
First, contrastive learning is performed using a large-scale, weakly-supervised dataset. 
The dataset used in the first stage typically consists of text pairs extracted from sources like Wikipedia and web corpora. 
Although this dataset is noisy and may contain false positives/negatives, large-scale training with substantial batch sizes has been shown to improve the quality of the embeddings. 
After contrastive pre-training, the model is fine-tuned using a manually labeled dataset. 
While the model from the first stage already yields reasonably effective embeddings, its performance can be further enhanced through fine-tuning with high-quality, human-labeled data.

Building on these methods, this report aims to develop a robust base model that can adapt to various domains through contrastive pre-training for Japanese.
However, unlike in English or multilingual models, there are several challenges in developing a general text embedding model for Japanese. 
One of the most significant challenges is the limited availability of training datasets. 
To address this, in addition to collecting and preprocessing existing datasets, we synthesized additional training data using Large Language Models (LLMs) for contrastive pre-training. 
In this section, we provide a detailed explanation of our contrastive pre-training process.

\subsection{Existing Datasets}
First, we collected and preprocessed available open datasets suitable for training text embedding models.
Specifically, we standardized the format and applied common preprocessing steps to 7 datasets:
Japanese Wikipedia\footlink{https://huggingface.co/datasets/hpprc/jawiki},
WikiBooks\footlink{https://huggingface.co/datasets/hpprc/jawiki-books},
Wiktionary\footlink{https://huggingface.co/datasets/hpprc/jawiki-wiktionary},
Japanese split of MQA\footlink{https://huggingface.co/datasets/clips/mqa},
Japanese split of CC News\footlink{https://huggingface.co/datasets/intfloat/multilingual_cc_news},
Japanese Research Corpus (JRC)\footlink{https://huggingface.co/datasets/kunishou/J-ResearchCorpus}\footsep\footnote{
Japanese Research Corpus is a high-quality dataset comprising Japanese academic papers.
While it includes papers from various academic societies, it also contains papers from the journal ``Natural Language Processing,'' which is included in the evaluation benchmark, JMTEB~\cite{JMTEB}.
To prevent potential leakage, we excluded the data from ``Natural Language Processing'' from our training dataset.
},
Wiki Atomic Edits~\cite{WikiAtomicEdits}.
We applied NFKC normalization and removed invisible characters.
Each dataset consists of pairs of ``queries'' typically composed of article titles, and ``passages'' composed of article bodies or longer texts.

\begin{table*}[t!]
\small
\centering
\begin{tabular}{llll@{\hspace{4ex}}r}
\bhline
\tabH Source & Anchor & Positive & Negative & Dataset size\\
\bhline
\tabH Wikipedia (1) & title + section title & 1-paragraph & random 1-paragraph & 19,361,464\\
Wikipedia (3) & title + section title & 3-paragraphs & random 3-paragraphs & 10,010,462\\
Wikipedia (long)  & title / abst. & abst. / article body & random abst. / article body & 7,889,486\\
Wiktionary & title & article body & random article body & 697,405\\
WikiBooks  & title + section title & 1-paragraph & random 1-paragraph & 314,207\\
\hline
\tabH MQA & title & article body & BM25 mined article body& 25,165,824\\
\hline
\tabH CC News (long)  & title & article body & BM25 mined article body& 6,248,336\\
CC News (short) & random sentence & sentence in the same article & sentence in other articles & 2,795,632\\
\hline
\tabH AutoWikiQA (MX)  & question & passage & BM25 mined passage& 11,563,562\\
AutoWikiQA (Nemo)  & question & passage & BM25 mined passage & 495,062\\
\hline
\tabH JRC & title + section title & section body & BM25 mined section body & 131,072\\
\hline
\tabH Wiki Atomic Edits & sentence & edited sentence & random sentence & 3,679,939\\
AutoWikiNLI   & premise & hypothesis (entailment) & hypothesis (contradiction) & 203,147\\
JSNLI  & premise & hypothesis (entailment) & hypothesis (contradiction) & 180,146\\
\hline
\tabH Total & & &  &  88,735,744\\
\bhline
\end{tabular}
\vspace{-1ex}
\caption{
Datasets used for contrastive pre-training
}
\label{tab:contrastive-pretraining-datasets}
\end{table*}

\begin{table*}[t!]
\small
\centering
\begin{tabular}{lccccc}
\bhline
\tabH Model & \#Params.  & GPUs & Base LM  \\
\bhline
\tabH \RuriPTSmall ~(\hf{cl-nagoya/ruri-pt-small}) & \ \ 68M & A6000$\times$4 & \hf{line-corporation/line-distilbert-base-japanese} \\
\RuriPTBase ~(\hf{cl-nagoya/ruri-pt-base}) & 111M & A100$\times$4 & \hf{tohoku-nlp/bert-base-japanese-v3} \\
\RuriPTLarge ~(\hf{cl-nagoya/ruri-pt-large}) & 337M & A100$\times$4 & \hf{tohoku-nlp/bert-large-japanese-v2} \\
\bhline
\end{tabular}
\vspace{-1ex}
\caption{
Overview of the model with contrastive pre-training.
}
\label{tab:model-summary-pt}
\end{table*}

\subsection{Synthesized Datasets}

The use of synthetic datasets in training text embeddings is very promising and is actively explored~\cite{SynCSE,MistralE5,SomaSato,Gecko}.
This is particularly true for languages like Japanese, where there are few available datasets for training embedding models, and licensing is crucial.

Therefore, we created synthetic datasets using LLMs for two types of datasets commonly used in embedding model training: QA datasets and natural language inference (NLI) datasets, and used them for model training.
While the synthetic datasets are considered relatively high quality, they may contain noise or bias. Thus, we included them in the pretraining datasets.

\begin{description}[leftmargin=0.5cm]
\item[AutoWikiQA\footnotemark]{
\foothreftext{https://huggingface.co/datasets/cl-nagoya/auto-wiki-qa}{https://huggingface.co/datasets/cl-nagoya/auto-wiki-qa}
is a dataset consisting of queries and answers generated from random Wikipedia passages.
For generation, we mainly used Swallow-MX\footlink{https://huggingface.co/tokyotech-llm/Swallow-MX-8x7b-NVE-v0.1}, which is a continually pre-trained model from Mixtral-8x7B\footlink{https://huggingface.co/mistralai/Mixtral-8x7B-Instruct-v0.1} on the cleaned large Japanese corpus~\cite{Swallow-Corpus} as well as Swallow~\cite{Swallow}.
We also used Nemotron-4 340B~\cite{Nemotron}.
The source passages to generate queries and answers were constructed by concatenating three paragraphs from random Wikipedia articles to ensure a sufficient length of the text.
Since the passages in AutoWikiQA are created by combining multiple paragraphs from Wikipedia, a single passage usually consists of multiple sentences rather than just one.
The resulting dataset consists of over 250 million query--passage pairs.
}

\item[AutoWikiNLI\footnotemark]{
\foothreftext{https://huggingface.co/datasets/cl-nagoya/auto-wiki-nli-reward}{https://huggingface.co/datasets/cl-nagoya/auto-wiki-nli-reward}
is a synthesized natural language inference (NLI) dataset generated using Nemotron-4 340B. 
We sampled random sentences from Wikipedia as premises and generated both entailment and contradiction sentences. 
By generating both entailment and contradiction sentences from a single premise, we create triplet datasets that include harder negatives, which are crucial for contrastive learning due to the lexical similarity between entailment and contradiction sentences.
Initially, we observed that when the LLM generated entailment followed by contradiction sentences, the contradiction sentences were often simple negations.
To improve this, we reversed the generation order, creating contradiction sentences first, followed by entailment sentences.
Additionally, some generated hypothesis sentences were of low quality.
To address this, we used the reward model of Nemotron-4 340B\footlink{https://build.nvidia.com/nvidia/nemotron-4-340b-reward} to score the generated outputs and removed the bottom 20\% of examples based on low helpfulness scores.
}

\end{description}

\subsection{Pre-training Dataset}

Table~\ref{tab:contrastive-pretraining-datasets} shows the datasets used for contrastive pre-training.
Our training strategy aims to achieve high-performance models by leveraging a diverse range of datasets, including both noisy and high-quality, manually curated sources.
This approach allows the model to learn robust representations from varied data before fine-tuning on more specific, high-quality datasets.
Wikipedia was utilized in multiple ways to create text pairs for training.
We employed various methods to extract and pair texts from Wikipedia, maximizing the value of this rich information source for our pre-training process.
Also, we incorporated JSNLI~\cite{JSNLI}, a dataset derived from machine translation of English~\cite{SNLI}.

We implemented hard negative mining during the pre-training phase, a technique shown to enhance model performance as reported in \citet{E5}.
Although \citet{E5} suggests that hard negative mining becomes impractical for datasets approaching 200 million samples, our relatively smaller dataset size allowed us to effectively employ this technique.
We utilized BM25 for generating hard negatives, which required preprocessing the entire document corpus to create searchable indexes.
To optimize this process, we created separate indexes for each dataset, thereby reducing the time and computational costs associated with indexing and searching.

\subsection{Training Details}

An overview of contrastive pre-trained model is shown in Table~\ref{tab:model-summary-pt}.
In the contrastive pre-training, we built three models of different sizes: small, base, and large.
The small model was based on \hf{line-corporation/line-distilbert-base-japanese}, the base model on \hf{tohoku-nlp/bert-base-japanese-v3}, and the large model on \hf{tohoku-nlp/bert-large-japanese-v2}, with contrastive pre-training applied as fine-tuning.
All of these models are Japanese BERT~\cite{BERT} models.
We used the improved contrastive loss, as proposed by~\citet{GTE}, for training.
This loss function not only calculates the similarity between query and passage in contrastive learning using in-batch negatives, but also considers the similarities between query--query, passage--query, and passage--passage.
It aims to lower the similarity scores for non-positive examples.
In this regard, it is similar to the loss function proposed by~\citet{PairSupCon}, which also emphasizes the use of in-batch negatives as much as possible.
For training the small model, we used four NVIDIA A6000 GPUs, while for the base and large models, we used four NVIDIA A100 (80GB) GPUs.

\paragraph{Prefix}
Recent embedding models commonly use prefixes in addition to the text being embedded~\cite{E5,mE5,GTE,BGE,MistralE5}.
This approach is known to be particularly effective for tasks requiring asymmetric similarity, such as retrieval tasks.
Therefore, we also utilized prefixes during model training. Specifically, we added the prefix ``クエリ: '' to search queries and ``文章: '' to target passages.
While English or multilingual models often use ``query: '' for queries and ``passage: '' for passages, since our model is a Japanese model, we simply translated these prefixes into Japanese.

\paragraph{Batching Strategy}
As with GTE~\cite{GTE} and InstructOR~\cite{InstructOR}, only triplets from the same dataset were included in a single batch to prevent shortcut learning.
This batching strategy is called task-homogeneous batching.
Task-homogeneous batching has another advantage: it prevents the mixing of datasets with different sequence lengths in the same batch, reducing the need for padding tokens and thus decreasing training time.
To further improve peformance, if identical sentences are mixed within the same batch, false negatives can occur, we removed duplicate sentences within each batch in advance.

\paragraph{Hyperparameters and Implementations}

Following \citet{GTE}, which reported minimal performance improvements for batch sizes larger than 8192, we set the batch size to 8192 for the contrastive pre-training phase.
As with E5, the positional embedding was fixed\footlink{https://github.com/microsoft/unilm/issues/1120}.
Unlike SimCSE, but similar to SimLM~\cite{SimLM}\footlink{https://github.com/microsoft/unilm/blob/9c0f1ff7ca53431fe47d2637dfe253643d94185b/simlm/src/config.py\#L54} and E5, we did not use a pooler layer.
We performed data augmentation by shuffling the sentence order of positive example documents.
For splitting sentences, we used Konoha\footlink{https://github.com/himkt/konoha}.
For other hyperparameters, refer to the left side of Table~\ref{tab:hyperparameters-emb}.

\section{Building Reranker}
\label{sec:reranker}

A reranker is a model typically used in retrieval tasks, where the query and document are concatenated and input into the model to output a relevance score.
Unlike dual-encoder, which independently embeds the text and measures similarity in vector space, the reranker---also known as a cross-encoder---captures the interaction between the query and the document.
This allows it to measure relevance more accurately than a dual-encoder~\cite{SimLM}.
Recent embedding models have found it effective to incorporate knowledge distillation from cross-encoders in addition to contrastive learning~\cite{SimLM,E5}.
Knowledge distillation from cross-encoders is a method where the model is trained to align the similarity score distributions between query and document produced by the cross-encoder and those produced by the dual-encoder.

In the model constructed in this report, we follow E5 and apply knowledge distillation from cross-encoders.
To achieve this, we first built a reranker for Japanese.
The biggest challenge in constructing a Japanese reranker is the dataset.
Japanese retrieval datasets are extremely limited in size, making it difficult to prepare training datasets on the scale of those used in English or multilingual models.
Therefore, we adopted a two-stage learning approach, starting with training on noisy datasets, including the synthesized dataset from Section \ref{sec:contrastive-pre-training}, and then fine-tuning on higher-quality datasets.

\begin{table}[t]
\small
\centering
\begin{tabular}{lc}
\bhline
\tabH Source & Dataset size\\
\bhline
\tabH JSQuAD & 212,352\\
AutoWikiQA (Nemo) & 190,743\\
JaQuAD & 108,068\\
Quiz No Mori & \ \ 36,120\\
Quiz Works & \ \ 29,112\\
JQaRA & \ \ 16,260\\
MIRACL & \ \ 13,968\\
Mr. TyDi & \ \ \ \ 7,394\\
MKQA & \ \ \ \ 6,636\\
\hline
\tabH Total & 620,653\\
\bhline
\end{tabular}
\vspace{-1ex}
\caption{
Datasets used for training the reranker in the first stage.
}
\label{tab:reranker-stage1-datasets}
\end{table}

\begin{table}[t]
\small
\centering
\begin{tabular}{lc}
\bhline
\tabH Source & Dataset size\\
\bhline
\tabH Quiz No Mori & 18,060\\
Quiz Works & 14,556\\
JQaRA & \ \ 8,130\\
MIRACL & \ \ 6,984\\
MR. TyDi & \ \ 3,697\\
\hline
\tabH Total & 51,427 \\
\bhline
\end{tabular}
\vspace{-1ex}
\caption{
Datasets used for training the reranker in the second stage.
}
\label{tab:reranker-stage2-datasets}
\end{table}

\subsection{Datasets}

We trained the reranker in two stages: the first stage used a large but noisy dataset, while the second stage employed a smaller, higher-quality dataset along with JaColBERT v2.5~\cite{JaColBERTv2.5}.
The datasets used in the first stage are listed in Table~\ref{tab:reranker-stage1-datasets}, and those used in the second stage are shown in Table~\ref{tab:reranker-stage2-datasets}\footnote{
We did not use the MMARCO~\cite{MMARCO} dataset for our retrieval tasks.
MMARCO is a translated version of MS MARCO~\cite{MSMARCO}, which is primarily intended for non-commercial research purposes.
There were licensing concerns when publishing models with a commercial-use license.
Additionally, the quality of the Japanese translations in MMARCO is relatively low, and in preliminary experiments, training with MMARCO negatively affected the performance of the reranker.
}.
For reranker training, we utilized existing Japanese retrieval and QA datasets, including JSQuAD~\cite{JGLUE}, JaQuAD~\cite{JaQuAD}, JQaRA~\cite{JQaRA}, MKQA~\cite{MKQA}, Mr. TyDi~\cite{MrTyDi}, and MIRACL~\cite{MIRACL}.
We also used a synthesized dataset generated by Nemotron-4 340B~\cite{Nemotron}.
Additionally, we used high-quality QA datasets extracted from Japanese quiz websites, available under a free license\footlink{https://huggingface.co/datasets/hpprc/quiz-works}\footsep\footlink{https://huggingface.co/datasets/hpprc/quiz-no-mori}.
For each dataset, we used only the train set for those with predefined splits, and for others, we used all available data.
As a result, we did not use the examples included in the test sets of the benchmarks for training.

\paragraph{Pseudo Positives and Hard Negative Mining}

To train rerankers, each query is paired with one positive document and multiple negative documents.
The model is trained to ensure that the relevance score for the query--positive document pair is higher than for any query--negative document pair.
There are two key points in reranker training; the first is to use challenging examples as hard negatives, and the second is to avoid false negatives, where a negative document is actually a positive example.

To collect hard negatives for each query, we used a combination of BM25-based negative mining and nearest neighbor search based on the embeddings of the multilingual E5-large model~\cite{mE5} (mE5-large).
Specifically, we integrated the results of BM25 and mE5-large negative mining using reciprocal rank fusion (RRF)~\cite{RRF}, and selected hard negatives from the examples excluding the top-ranked negatives.
We used documents ranked between 30th and 100th by the combined BM25 and mE5-large ranking as hard negatives.
To mitigate the issue of false negatives, we used the answers from QA datasets.
Specifically, after hard negative mining, we removed documents containing the query's answer as pseudo positives and used them as ``mined positives'' as well as JAQKET~\cite{JAQKET}.

For hard negative mining, we used a collection of concatenated 3-paragraphs from Japanese Wikipedia, Wiktionary, and WikiBooks, for the Japanese QA datasets as candidates for negatives excluding Mr. TyDi and MIRACL.
For Mr. TyDi and MIRACL, we used the predefined document collections as they were.

\subsection{Training Details}

We trained the reranker in two-stages.
In the first stage, we built a reranker using noisy data, and in the second stage, we fine-tuned the reranker.
For the first stage, we applied data augmentation by shuffling the sentence order of each positive document in the datasets, but no shuffling was done in the second stage.
The sequence length during training was set to a relatively short value of 256 in the first stage and increased to 512 in the second stage.
The number of hard negatives used during training was set to 63 both for the first and second stage.
As the loss function, we used cross-entropy loss, where the score of the positive document is maximized relative to the scores of the 63 negative documents for each query.
We used the contrastive pre-trained model constructed in Section~\ref{sec:contrastive-pre-training} as the base model.
For other hyperparameters, refer to Table~\ref{tab:hyperparameters-reranker}.

\begin{table*}[t]
\small
\centering
\begin{tabular}{lcccc}
\bhline
\tabH Model & \#Param. (w/o Emb.) & JQaRA & JaCWIR & MIRACL\\
\bhline
\tabH \hf{hotchpotch/japanese-reranker-cross-encoder-xsmall-v1} & 107M \ \ \ (11M) & 61.4 & 93.8 & 90.6 \\
\hf{hotchpotch/japanese-reranker-cross-encoder-small-v1} & 118M \ \ \ (21M) & 62.5 & 93.9 & 92.2 \\
\hf{hotchpotch/japanese-reranker-cross-encoder-base-v1} & 111M \ \ \ (86M) & 67.1 & 93.4 & 93.3 \\
\hf{hotchpotch/japanese-reranker-cross-encoder-large-v1} & 337M \ (303M) & 71.0 & 93.6 & 91.5 \\
\hline
\tabH \hf{hotchpotch/japanese-bge-reranker-v2-m3-v1} & 568M \ (303M) & 69.2 & 93.7 & 94.7 \\
\hf{BAAI/bge-reranker-v2-m3} & 568M \ (303M) & 67.3 & 93.4 & 94.9 \\
\hline
\tabH \RuriRerankerSmall ~(\hf{cl-nagoya/ruri-reranker-small}) & \ \ 68M \ \ \ (43M)\  & 64.5 & 92.6 & 92.3 \\
\RuriRerankerBase ~(\hf{cl-nagoya/ruri-reranker-base}) & 111M \ \ \ (86M) & 74.3 & 93.5 & 95.6 \\
\RuriRerankerLarge ~(\hf{cl-nagoya/ruri-reranker-large}) & 337M \ (303M) & \textbf{77.1} & \textbf{94.1} & \textbf{96.1} \\
\bhline
\end{tabular}
\vspace{-1ex}
\caption{
Performance comparison of rerankers.
JQaRA is evaluated using nDCG@10, JaCWIR with MAP@10, and MIRACL with Recall@30. ``\#Param. (w/o Emb.)'' indicates the number of parameters, both with and without token embeddings.
}
\label{tab:reranker-result}
\end{table*}

\subsection{Evaluation}

\paragraph{Settings}
The reranker is used for dataset filtering and knowledge distillation for embedding models, so its performance is expected to impact the performance of resulting embedding models.
Therefore, we first evaluated the reranker for Japanese.
In reranker evaluation, the goal is to determine how well the model ranks the relevant documents at the top when given a query and a set of documents.
We used JQaRA~\cite{JQaRA}, JaCWIR~\cite{JaCWIR}, and the test set of MIRACL~\cite{MIRACL} for evaluation.
JQaRA is a dataset designed to evaluate the retrieval of useful data for answering questions, which is important for retrieval-augmented generation (RAG).
JaCWIR is a diverse retrieval evaluation dataset based on web articles.
For the evaluation of JQaRA and JaCWIR, we used the official evaluation code\footlink{https://github.com/hotchpotch/JQaRA}\footsep\footlink{https://github.com/hotchpotch/JaCWIR}, and for MIRACL, we used an implementation designed for reranking with mined negatives\footlink{https://github.com/oshizo/JapaneseEmbeddingEval}.
The evaluation metrics were top-10 nDCG (nDCG@10) for JQaRA, top-10 mean average precision (MAP@10) for JaCWIR, and top-30 recall (Recall@30) for MIRACL.

\paragraph{Results}
Table~\ref{tab:reranker-result} shows the evaluation results of major multilingual and Japanese rerankers, as well as our reranker (Ruri-Reranker).
Ruri-reranker demonstrated consistently strong performance across the board, with the base model achieving performance comparable to or exceeding existing Japanese and multilingual rerankers, and the large model significantly outperforming them.
Notably, our model performed particularly well on JQaRA, which can be attributed to the use of diverse QA datasets during training.

\begin{table}[t]
\small
\tabcolsep 4pt
\centering
\begin{tabular}{lcccc}
\bhline
\tabH Model & Stage & JQaRA & JaCWIR & MIRACL\\
\bhline
\tabH \RuriPTSmall & 1 only & 63.9 & 92.5 & 91.2 \\
\RuriPTSmall & 2 only & 60.3 & 89.9 & 89.3 \\
\RuriPTSmall & 1 $\rightarrow$ 2 & \textbf{64.5} & \textbf{92.6} & \textbf{92.3} \\
\hline
\tabH \RuriPTBase & 1 only & 72.9 & 92.4 & 94.2 \\
\RuriPTBase & 2 only & 68.0 & 92.9 & 93.7 \\
\RuriPTBase & 1 $\rightarrow$ 2 & \textbf{74.3} & \textbf{93.5} & \textbf{95.6} \\
\hline
\tabH \RuriPTLarge & 1 only & 75.8 & 93.4 & 95.4 \\
\RuriPTLarge & 2 only & 70.5 & 90.8 & 93.2 \\
\RuriPTLarge & 1 $\rightarrow$ 2 & \textbf{77.1} & \textbf{94.1} & \textbf{96.1} \\
\bhline
\end{tabular}
\vspace{-1ex}
\caption{
The results of the ablation study for two-stage training.
The evaluation metrics are the same as those in Table~\ref{tab:reranker-result}.
}
\label{tab:reranker-ablation-stage}
\end{table}

\begin{table}[t]
\small
\centering
\tabcolsep 4pt
\begin{tabular}{lcccc}
\bhline
\tabH Model & Phase & JQaRA & JaCWIR & MIRACL\\
\bhline
\tabH $\text{BERT}_\text{small}$ & stage1 & 63.7 & 89.4 & 90.4 \\
$\text{BERT}_\text{small}$ & stage2 & 64.3 & 91.4 & 91.6 \\
\RuriPTSmall & stage1 & 63.9 & 92.5 & 91.2 \\
\RuriPTSmall & stage2 & \textbf{64.5} & \textbf{92.6} & \textbf{92.3} \\
\hline
\tabH $\text{BERT}_\text{base}$ & stage1 & 71.8 & 89.3 & 93.9 \\
$\text{BERT}_\text{base}$ & stage2 & 73.1 & 91.6 & 95.1 \\
\RuriPTBase & stage1 & 72.9  & 92.4 & 94.2 \\
\RuriPTBase & stage2 & \textbf{74.3} & \textbf{93.5} & \textbf{95.6 }\\
\hline
\tabH $\text{BERT}_\text{large}$ & stage1 & 76.1 & 92.2 & 95.2 \\
$\text{BERT}_\text{large}$ & stage2 & \textbf{77.3} & 93.5 & 96.0 \\
\RuriPTLarge & stage1 & 75.8 & 93.4 & 95.4 \\
\RuriPTLarge & stage2 & 77.1 & \textbf{94.1} & \textbf{96.1} \\
\bhline
\end{tabular}
\vspace{-1ex}
\caption{
The results of the ablation study for the benefit of contrastive pre-training.
The evaluation metrics are the same as those in Table~\ref{tab:reranker-result}.
}
\label{tab:reranker-ablation-pt}
\end{table}

\paragraph{Ablation Study}
We conducted several ablation studies on the design of our reranker.
The two main points we investigated were: 1) whether the two-stage training of the reranker is effective, and 2) whether using a contrastive pre-trained model as the base model for the reranker is beneficial.

First, for the two-stage reranker training, we experimented with three configurations: 1) training only the first stage, 2) training only the second stage, and 3) training from the first stage to the second stage (i.e. Ruri-Reranker).
Table~\ref{tab:reranker-ablation-stage} shows the results.
From the table, it is clear that two-stage training consistently yielded the best performance across all model sizes.
While training only in the first stage achieved reasonable performance, adding fine-tuning with higher-quality datasets in the second stage further improved the results, demonstrating the effectiveness of the two-stage approach.
This observation is consistent with \citet{JaColBERTv2.5}.

Next, we investigated whether using a contrastive pre-trained model (Ruri-PT) as the base model for the reranker is beneficial.
Specifically, we compared the performance of a contrastive pre-trained model and a non-pre-trained model, both fine-tuned as rerankers in the same manner.
Table~\ref{tab:reranker-ablation-pt} shows the results.
Observing the models and stages, it is evident that the Ruri-PT generally outperformed the non-pre-trained model.
While the training objective of contrastive pre-training differs from that of reranking, this result suggests that contrastive pre-training helps the model learn how to focus on important information in the text, leading to improved performance.

\section{Supervised Fine-tuning}

Following previous research, we constructed the final embedding model by fine-tuning a contrastive pre-trained model, trained on a weakly supervised dataset, using high-quality datasets. This section describes the datasets used for training, details of the training process, and evaluation.

\subsection{Dataset}

For fine-tuning the contrastive pre-trained model, we collected high-quality datasets.
The datasets are shown in Table~\ref{tab:fine-tuning-datasets}.
Our datasets can be categorized into two types: retrieval/QA datasets and natural language inference (NLI) datasets.
The retrieval/QA datasets were the same as those used in the second stage of training, described in Section~\ref{sec:reranker}.
For the NLI datasets, we used NU-SNLI\footlink{https://huggingface.co/datasets/cl-nagoya/nu-snli} and NU-MNLI\footlink{https://huggingface.co/datasets/cl-nagoya/nu-mnli}, which were translated from SNLI~\cite{SNLI} and MNLI~\cite{MNLI} using a model fine-tuned on machine translation tasks with Swallow-MX.
While NU-SNLI and NU-MNLI were not manually created, their quality is sufficiently high.
We also used JaNLI~\cite{JaNLI} to facilitate the model in capturing more subtle differences in meaning.

To perform knowledge distillation from the cross-encoder, we used \RuriRerankerLarge, constructed in Section~\ref{sec:reranker}, to score the relevance of the retrieval datasets.
Additionally, to mitigate the negative effect from noisy examples in the retrieval datasets, we removed examples where the relevance score between the query and the positive document was less than 0.8.
As data augmentation, we also added examples where the sentence order of positive documents was shuffled, along with the original, unshuffled examples.
For the NLI datasets, we did not apply knowledge distillation, so no relevance scoring by the reranker was performed.
As in Section~\ref{sec:contrastive-pre-training}, we used task-homogeneous batching to ensure that only examples from the same dataset were included in each batch.

\begin{table}[t]
\small
\centering
\begin{tabular}{lcc}
\bhline
\tabH Source & Distill. & Dataset size\\
\bhline
\tabH Quiz No Mori & \checkmark & \ \ 31,232\\
Quiz Works & \checkmark & \ \ 26,624\\
JQaRA & \checkmark & \ \ 13,824\\
MIRACL & \checkmark & \ \ 12,800\\
Mr. TyDi & \checkmark & \ \ \ \ 7,168\\
\hline
\tabH NU-SNLI & & 109,568\\
NU-MNLI & & \ \ 77,824\\
JaNLI & & \ \ 13,824\\
\hline
\tabH Total & & 292,864\\
\bhline
\end{tabular}
\vspace{-1ex}
\caption{
Datasets used for supervised fine-tuning.
}
\label{tab:fine-tuning-datasets}
\end{table}

\begin{table*}[t!]
\small
\centering
\begin{tabular}{lccccccc}
\bhline
\tabH Model & \#Params. & Dim. & \#Layer & Pooling & Context Len. & Vocab Size & JMTEB Avg. \\
\bhline
\tabH \RuriSmall ~(\hf{cl-nagoya/ruri-small}) & \ \ 68M & \ \ 768 & \ \ 6 & Mean & 512 & 32,768 & 71.53 \\
\RuriBase ~(\hf{cl-nagoya/ruri-base}) & 111M & \ \ 768 & 12 & Mean & 512 & 32,768 & 71.91 \\
\RuriLarge ~(\hf{cl-nagoya/ruri-large}) & 337M & 1024 & 24 &Mean & 512 & 32,768 & 73.31 \\
\bhline
\end{tabular}
\vspace{-1ex}
\caption{
Overview of the model with supervised fine-tuning.
}
\label{tab:model-summary-ft}
\end{table*}

\begin{table*}[t!]
\small
\centering
\begin{tabular}{@{\ \ }lcccccccc@{\ \ }}
\bhline
\tabH Model & \#Param. & Retrieval & STS & Class. & Reranking & Clustering & Pair. & Avg. \\
\bhline
\tabH \hf{cl-nagoya/sup-simcse-ja-base} & 111M & 49.64 & 82.05 & 73.47 & 91.83 & 51.79 & 62.57 & 63.36\\
\hf{cl-nagoya/sup-simcse-ja-large} & 337M & 37.62 & 83.18 & 73.73 & 91.48 & 50.56 & 62.51 & 58.88\\
\hf{cl-nagoya/unsup-simcse-ja-base} & 111M & 40.23 & 78.72 & 73.07 & 91.16 & 44.77 & 62.44 & 58.39\\
\hf{cl-nagoya/unsup-simcse-ja-large} & 337M & 40.53 & 80.56 & 74.66 & 90.95 & 48.41 & 62.49 & 59.58\\
\hf{pkshatech/GLuCoSE-base-ja} & 133M & 59.02 & 78.71 & 76.82 & 91.90 & 49.78 & 66.39 & 67.29\\
\hline
\tabH \hf{sentence-transformers/LaBSE} & 472M & 40.12 & 76.56 & 72.66 & 91.63 & 44.88 & 62.33 & 58.01\\
\hf{intfloat/multilingual-e5-small} & 118M & 67.27 & 80.07 & 67.62 & 93.03 & 46.91 & 62.19 & 67.71\\
\hf{intfloat/multilingual-e5-base} & 278M & 68.21 & 79.84 & 69.30 & 92.85 & 48.26 & 62.26 & 68.61 \\
\hf{intfloat/multilingual-e5-large} & 560M & 70.98 & 79.70 & 72.89 & 92.96 & 51.24 & 62.15 & 70.90\\
\hline
\tabH OpenAI/text-embedding-ada-002 & - & 64.38 & 79.02 & 69.75 & 93.04 & 48.30 & 62.40 & 67.21 \\
OpenAI/text-embedding-3-small & - & 66.39 & 79.46 & 73.06 & 92.92 & 51.06 & 62.27 & 69.18 \\
OpenAI/text-embedding-3-large & - & 74.48 & 82.52 & 77.58 & 93.58 & 53.32 & 62.35 & 74.05 \\
\hline
\tabH \RuriSmall ~(\hf{cl-nagoya/ruri-small}) & \ \ 68M & 69.41 & 82.79 & 76.22 & 93.00 & 51.19 & 62.11 & 71.53\\
\RuriBase ~(\hf{cl-nagoya/ruri-base}) & 111M & 69.82 & 82.87 & 75.58 & 92.91 & 54.16 & 62.38 & 71.91\\
\RuriLarge ~(\hf{cl-nagoya/ruri-large}) & 337M & 73.02 & 83.13 & 77.43 & 92.99 & 51.82 & 62.29 & 73.31\\
\bhline
\end{tabular}
\vspace{-1ex}
\caption{
Evaluation results on JMTEB. 
``\#Param.'' indicates the number of model parameters, 
``Retrieval'' shows the average performance on 6 retrieval datasets, 
``STS'' on 2 Semantic Textual Similarity datasets, 
``Classification'' on 4 classification datasets, 
``Reranking'' on 1 reranking dataset, 
``Clustering'' on 2 clustering datasets, 
``Pair.'' on 1 pair classification dataset, 
and ``Avg.'' is the micro average across all 16 datasets.
}
\label{tab:main-result}
\end{table*}

\begin{table*}[t!]
\small
\tabcolsep 4.5pt
\centering
\begin{tabular}{lccccccc}
\bhline
\tabH \multirow{2}{*}{Model} & \multirow{2}{*}{JaGovFAQs} & \multirow{2}{*}{JAQKET} & \multirow{2}{*}{Mr. TyDi} & \multicolumn{3}{c}{NLP Journal} & \multirow{2}{*}{Avg.} \\
 &  &  &  & Abst.--Intro. & Title--Abst. & Title--Intro. &  \\
\bhline
\tabH \hf{pkshatech/GLuCoSE-base-ja} & 63.88 & 39.82 & 30.28 & 78.26 & 82.06 & 59.82 & 59.02\\
\hline
\tabH \hf{intfloat/multilingual-e5-small} & 64.11 & 49.97 & 36.05 & 85.21 & 95.26 & 72.99 & 67.27\\
\hf{intfloat/multilingual-e5-base} & 65.34 & 50.67 & 38.38 & 87.10 & 94.73 & 73.05 & 68.21 \\
\hf{intfloat/multilingual-e5-large} & 70.30 & 58.78 & \textbf{43.63} & 86.00 & 94.70 & 72.48 & 70.98 \\
\hline
\tabH OpenAI/text-embedding-ada-002 & 61.02 & 42.56 & 14.51 & 94.99 & 91.23 & 81.98 & 64.38\\
OpenAI/text-embedding-3-small & 64.02 & 33.94 & 20.03 & 98.47 & 91.70 & 90.17 & 66.39\\
OpenAI/text-embedding-3-large & 72.41 & 48.21 & 34.88 & \textbf{99.33} & \textbf{96.55} & \textbf{95.47} & 74.48\\
\hline
\tabH \RuriSmall ~(\hf{cl-nagoya/ruri-small}) & 73.65 & 48.44 & 33.43 & 87.69 & 97.17 & 76.09 & 69.41\\
\RuriBase ~(\hf{cl-nagoya/ruri-base}) & 74.56 & 50.12 & 35.45 & 86.89 & \textbf{96.57} & 75.31 & 69.82\\
\RuriLarge ~(\hf{cl-nagoya/ruri-large}) & \textbf{76.68} & \textbf{61.74} & 38.03 & 87.12 & \textbf{96.58} & 77.97 & 73.02\\
\bhline
\end{tabular}
\vspace{-1ex}
\caption{
Evaluation results on the retrieval tasks.
We used nDCG@10 as an evaluation metric for all tasks.
}
\label{tab:retrieval-result}
\end{table*}

\subsection{Training Details}

We built a high-performance embedding model by fine-tuning the contrastive pre-trained model constructed in Section~\ref{sec:contrastive-pre-training} using high-quality datasets.
An overview of each model is shown in Table~\ref{tab:model-summary-ft}.
Following \citet{JaColBERTv2.5}, we decoupled the loss for knowledge distillation and contrastive learning.
Specifically, for retrieval/QA dataset examples, we computed the loss using knowledge distillation, and for NLI examples, we computed the loss using contrastive learning.

During knowledge distillation, inspired by \citet{JaColBERTv2.5}, we applied min-max normalization to both the student scores, calculated via cosine similarity of embeddings, and the teacher scores, calculated by the cross-encoder.
For the NLI dataset, we used the improved contrastive loss, as described in Section~\ref{sec:contrastive-pre-training}.
The maximum sequence length was set to 512, the batch size to 512, and the number of hard negatives to 15.
For other hyperparameters, refer to the right side of Table~\ref{tab:hyperparameters-emb}.

\subsection{Evaluation}
We evaluated our Japanese general text embedding model, Ruri, on the Japanese text embedding benchmark.

\paragraph{Settings}
For evaluation, we used JMTEB~\cite{JMTEB}, the Japanese version of the massive text embedding benchmark (MTEB)~\cite{MTEB}.
JMTEB includes 16 evaluation datasets covering various tasks such as classification, retrieval, and clustering.
We used the official implementation for the evaluation.

\paragraph{Results}
Table~\ref{tab:main-result} shows the results.
The results indicate that our model consistently outperforms existing multilingual embedding models such as mE5 and Japanese embedding models on average.
Notably, our base-sized model achieved higher average performance than the mE5-large.
Even when compared to proprietary embedding models, our model demonstrates comparable performance.

\begin{table*}[t!]
\small
\centering
\begin{tabular}{lccccccc}
\bhline
\tabH Model & Retrieval & STS & Class. & Reranking & Clustering & Pair. & Avg. \\
\bhline
\tabH \RuriPTLarge & \textbf{71.48} & 82.06 & 76.12 & \textbf{92.75} & \textbf{53.41} & 62.27 & \textbf{72.46}\\
\RuriPTLarge ~w/o retrieval & 68.08 & \textbf{82.32} & \textbf{76.42} & 92.66 & 51.98 & 62.29 & 71.11\\
\bhline
\end{tabular}
\vspace{-1ex}
\caption{
Performance of pre-trained models on JMTEB with and without using synthesized retrieval datasets.
}
\label{tab:ablation-synthesized-dataset}
\end{table*}

\begin{table*}[t!]
\small
\centering
\begin{tabular}{lccccccc}
\bhline
\tabH Model & Retrieval & STS & Class. & Reranking & Clustering & Pair. & Avg. \\
\bhline
\tabH \RuriPTSmall & 67.39 & 81.41 & 75.41 & 92.98 & 51.13 & 62.44 & 70.41\\
\RuriSmall ~w/o pre-training & 56.62 & 82.45 & 77.30 & 92.01 & 47.77 & 62.42 & 66.49\\
\RuriSmall & 69.41 & 82.79 & 76.22 & 93.00 & 51.19 & 62.11 & \textbf{71.53}\\
\hline
\tabH \RuriPTBase & 68.18 & 81.81 & 74.56 & 92.82 & 53.35 & 62.33 & 70.80\\
\RuriBase ~w/o pre-training & 52.99 & 81.95 & 76.19 & 91.60 & 51.85 & 62.20 & 65.25\\
\RuriBase & 69.82 & 82.87 & 75.58 & 92.91 & 54.16 & 62.38 & \textbf{71.91}\\
\hline
\tabH \RuriPTLarge & 71.48 & 82.06 & 76.12 & 92.75 & 53.41 & 62.27 & 72.46\\
\RuriLarge ~w/o pre-training & 57.84 & 83.66 & 76.50 & 91.51 & 49.56 & 62.35 & 67.09\\
\RuriLarge & 73.02 & 83.13 & 77.43 & 92.99 & 51.82 & 62.29 & \textbf{73.31}\\
\bhline
\end{tabular}
\vspace{-1ex}
\caption{
Evaluation results of the model with and without contrastive pre-training.
``w/o pre-training'' represents the performance of the model that underwent only supervised fine-tuning without contrastive pre-training.
}
\label{tab:ablation-pre-training}
\end{table*}

To further analyze the trends of each model, we focused on retrieval tasks, where performance differences tend to be more pronounced.
Table~\ref{tab:retrieval-result} shows the performance of each model on these tasks.
JaGovFAQs, JAQKET, and Mr. TyDi are standard QA/retrieval tasks, while the three NLP Journal-related tasks involve retrieving abstracts or introductions based on paper titles, or retrieving introductions based on abstracts.
The results show that our model performs well on JaGovFAQs, a FAQ retrieval task, and JAQKET, a QA task\footnote{
We used JQaRA to train the reranker and fine-tuned models, and JQaRA shares some data with JAQKET.
However, to avoid using the JAQKET test set and prevent data leakage, we only used a portion of the JQaRA data (dev, unused split) for training.
}, with our model outperforming proprietary embeddings on average.
On the other hand, proprietary embedding models performed exceptionally well on the NLP Journal-related tasks.
Although the training data and methods for proprietary embeddings are not disclosed, these NLP Journal tasks can be viewed as topic similarity search tasks in LaTeX documents, suggesting that these models may have been largely trained on LaTeX documents.

\paragraph{Ablation Study}

The key differences between our embedding model and existing Japanese models are: 1) the use of a synthesized dataset for model training, and 2) the application of contrastive pre-training.
Therefore, we conducted an ablation study on these aspects.

First, Table~\ref{tab:ablation-synthesized-dataset} shows the performance of the large model on JMTEB when contrastive pre-training is performed with or without the synthesized retrieval dataset.
The results indicate a significant performance improvement in retrieval tasks when the synthesized retrieval dataset is used.
On the other hand, when the synthesized retrieval dataset is excluded, there is a slight improvement in STS and classification tasks, suggesting that the introduction of synthesized datasets may not be beneficial for all tasks.
However, since retrieval tasks are generally more challenging compared to other tasks, improving performance in these tasks by using synthesized datasets is valuable.

Next, to verify the effectiveness of contrastive pre-training under the assumption of supervised fine-tuning, we compared the performance of models after supervised fine-tuning, using both contrastive pre-trained models and non-pre-trained models as the base models.
Table~\ref{tab:ablation-pre-training} shows the results.
We can clearly observe that the presence or absence of contrastive pre-training has a significant impact on post-supervised fine-tuning performance.
The improvement in retrieval task performance is particularly notable, indicating the importance of contrastive pre-training for retrieval tasks.

\section{Conclusion and Future Work}

In this report, we described the process of building a general-purpose Japanese text embedding model, Ruri.
Our contributions can be summarized into the following five points:
\begin{enumerate}[leftmargin=0.5cm, itemsep=0.3pt]
    \item{We collected datasets for building a Japanese embedding model and made them public with a permissive license.}
    \item{To address the shortage of Japanese retrieval datasets, we constructed a synthesized dataset using LLMs and verified its effectiveness.}
    \item{We constructed a large-scale dataset for contrastive pre-training in Japanese and demonstrated its utility.}
    \item{We developed a reranker in Japanese, achieving the highest performance among existing Japanese and multilingual rerankers.}
    \item{We built a Japanese embedding model, Ruri, which significantly outperformed existing models on the Japanese text embedding benchmark.}
\end{enumerate}

While our model already demonstrates high performance, there are still many challenges left in the realm of Japanese text embedding.
Below, we outline some of the challenges and considerations that we were unable to address in this report.

\paragraph{Prefix}
Using more diverse prefixes, beyond the simple ``クエリ: '' and ``文章: '', could potentially improve overall performance.
Indeed, recent models employing instructions~\cite{MistralE5,LLM2Vec,InstructOR,NV-Embed} or more detailed prefixes~\cite{NOMIC}\footlink{https://huggingface.co/nomic-ai/nomic-embed-text-v1.5} have been developed.

\paragraph{Knowledge Distillation from Cross-Encoder}
Knowledge distillation from cross-encoders has been introduced in models like SimLM~\cite{SimLM} and E5~\cite{E5}, but recent models based on LLMs~\cite{LLM2Vec,NV-Embed} do not seem to incorporate knowledge distillation from cross-encoders.
Also, during the development of Ruri, we observed that introducing knowledge distillation made training slightly unstable.
Whether this technique is truly necessary remains an open question for further investigation.

\paragraph{Dataset}
The current dataset is still insufficient, especially in terms of web corpora compared to GLuCoSE\footlink{https://huggingface.co/pkshatech/GLuCoSE-base-ja} and other multilingual embedding models.
To build a general-purpose text embedding model usable in various domains, a more diverse and higher-quality dataset may be crucial.

\paragraph{Base LM and Pre-training}
There may be room for improvement in the performance of the Japanese BERT used as the base model.
It has been reported that even with the same training methods and datasets, the quality of text embeddings can vary greatly depending on the base model~\cite{JapaneseSimCSETechReport}.
Developing base models specifically pre-trained for embedding, such as SimLM~\cite{SimLM}, RetroMAE~\cite{RetroMAE}, and RetroMAE-2~\cite{RetroMAE2}, could potentially lead to even higher-performing embedding models.
Also, it should be noted that Ruri does not use code datasets, so it cannot be applied for code search.
To perform code search in a Japanese-specific model, it may be necessary to develop a bilingual model with both Japanese and English vocabularies, given that program code is primarily written in English.

\paragraph{Context Length}
The context length is short.
While recent large language models can process sequences as long as 32k tokens or longer, most embedding models can only handle around 512 tokens.
There are long-context embedding models, such as Jina BERT~\cite{JinaBERT}, which use ALiBi~\cite{ALiBi}, and it is important to develop robust text embedding models for longer sequences.
In particular, Japanese BERT does not incorporate architectural advancements used in recent LLMs, such as RoPE~\cite{RoPE} and SwiGLU~\cite{SwiGLU}.
By backporting these developments from LLM research, it may be possible to create models with longer context lengths and higher performance.

\paragraph{Evaluation}
The bias in both training and evaluation datasets is also a concern.
There are very few evaluation datasets based on web corpora for assessing Japanese text embeddings.
Although solving this issue is complicated by licensing and copyright constraints, it is a challenge that must be addressed to evaluate the model's broad applicability.

\vspace{1ex}
Research on Japanese text embeddings is still in its early stages.
We hope this report contributes to further progress in this field.


\bibliography{custom}

\begin{thebibliography}{46}
\expandafter\ifx\csname natexlab\endcsname\relax\def\natexlab#1{#1}\fi

\bibitem[{Bajaj et~al.(2018)Bajaj, Campos, Craswell, Deng, Gao, Liu, Majumder, McNamara, Mitra, Nguyen, Rosenberg, Song, Stoica, Tiwary, and Wang}]{MSMARCO}
Payal Bajaj, Daniel Campos, Nick Craswell, Li~Deng, Jianfeng Gao, Xiaodong Liu, Rangan Majumder, Andrew McNamara, Bhaskar Mitra, Tri Nguyen, Mir Rosenberg, Xia Song, Alina Stoica, Saurabh Tiwary, and Tong Wang. 2018.
\newblock \href {http://arxiv.org/abs/1611.09268} {{MS MARCO: A Human Generated MAchine Reading COmprehension Dataset}}.
\newblock \emph{arXiv:1611.09268}.

\bibitem[{BehnamGhader et~al.(2024)BehnamGhader, Adlakha, Mosbach, Bahdanau, Chapados, and Reddy}]{LLM2Vec}
Parishad BehnamGhader, Vaibhav Adlakha, Marius Mosbach, Dzmitry Bahdanau, Nicolas Chapados, and Siva Reddy. 2024.
\newblock \href {https://openreview.net/forum?id=IW1PR7vEBf} {{LLM2Vec: Large Language Models Are Secretly Powerful Text Encoders}}.
\newblock In \emph{First Conference on Language Modeling (COLM)}.

\bibitem[{Bonifacio et~al.(2021)Bonifacio, Jeronymo, Abonizio, Campiotti, Fadaee, Lotufo, and Nogueira}]{MMARCO}
Luiz Bonifacio, Vitor Jeronymo, Hugo~Queiroz Abonizio, Israel Campiotti, Marzieh Fadaee, Roberto Lotufo, and Rodrigo Nogueira. 2021.
\newblock \href {http://arxiv.org/abs/2108.13897} {{mMARCO: A Multilingual Version of the MS MARCO Passage Ranking Dataset}}.
\newblock \emph{arXiv:2108.13897}.

\bibitem[{Bowman et~al.(2015)Bowman, Angeli, Potts, and Manning}]{SNLI}
Samuel~R. Bowman, Gabor Angeli, Christopher Potts, and Christopher~D. Manning. 2015.
\newblock \href {https://doi.org/10.18653/v1/D15-1075} {{A large annotated corpus for learning natural language inference}}.
\newblock In \emph{Proceedings of the 2015 Conference on Empirical Methods in Natural Language Processing (EMNLP)}, pages 632--642.

\bibitem[{Clavié(2024)}]{JaColBERTv2.5}
Benjamin Clavié. 2024.
\newblock \href {http://arxiv.org/abs/2407.20750} {{JaColBERTv2.5: Optimising Multi-Vector Retrievers to Create State-of-the-Art Japanese Retrievers with Constrained Resources}}.
\newblock \emph{arXiv:2407.20750}.

\bibitem[{Cormack et~al.(2009)Cormack, Clarke, and B{\"u}ttcher}]{RRF}
Gordon~V. Cormack, Charles L.~A. Clarke, and Stefan B{\"u}ttcher. 2009.
\newblock \href {https://api.semanticscholar.org/CorpusID:12408211} {{Reciprocal rank fusion outperforms condorcet and individual rank learning methods}}.
\newblock In \emph{Proceedings of the 32nd international ACM SIGIR conference on Research and development in information retrieval (SIGIR)}.

\bibitem[{Devlin et~al.(2019)Devlin, Chang, Lee, and Toutanova}]{BERT}
Jacob Devlin, Ming-Wei Chang, Kenton Lee, and Kristina Toutanova. 2019.
\newblock \href {https://doi.org/10.18653/v1/N19-1423} {{BERT: Pre-training of Deep Bidirectional Transformers for Language Understanding}}.
\newblock In \emph{Proceedings of the 2019 Conference of the North {A}merican Chapter of the Association for Computational Linguistics: Human Language Technologies (NAACL)}, pages 4171--4186.

\bibitem[{Faruqui et~al.(2018)Faruqui, Pavlick, Tenney, and Das}]{WikiAtomicEdits}
Manaal Faruqui, Ellie Pavlick, Ian Tenney, and Dipanjan Das. 2018.
\newblock \href {https://doi.org/10.18653/v1/D18-1028} {{WikiAtomicEdits: A Multilingual Corpus of {W}ikipedia Edits for Modeling Language and Discourse}}.
\newblock In \emph{Proceedings of the 2018 Conference on Empirical Methods in Natural Language Processing (EMNLP 2018)}, pages 305--315.

\bibitem[{Fujii et~al.(2024)Fujii, Nakamura, Loem, Iida, Ohi, Hattori, Shota, Mizuki, Yokota, and Okazaki}]{Swallow}
Kazuki Fujii, Taishi Nakamura, Mengsay Loem, Hiroki Iida, Masanari Ohi, Kakeru Hattori, Hirai Shota, Sakae Mizuki, Rio Yokota, and Naoaki Okazaki. 2024.
\newblock \href {https://openreview.net/forum?id=TQdd1VhWbe} {{Continual Pre-Training for Cross-Lingual LLM Adaptation: Enhancing Japanese Language Capabilities}}.
\newblock In \emph{Proceedings of the First Conference on Language Modeling (COLM)}, COLM.

\bibitem[{Gao et~al.(2021)Gao, Yao, and Chen}]{SimCSE}
Tianyu Gao, Xingcheng Yao, and Danqi Chen. 2021.
\newblock \href {https://aclanthology.org/2021.emnlp-main.552} {{SimCSE: Simple Contrastive Learning of Sentence Embeddings}}.
\newblock In \emph{Proceedings of the 2021 Conference on Empirical Methods in Natural Language Processing (EMNLP)}, pages 6894--6910.

\bibitem[{Günther et~al.(2024)Günther, Ong, Mohr, Abdessalem, Abel, Akram, Guzman, Mastrapas, Sturua, Wang, Werk, Wang, and Xiao}]{JinaBERT}
Michael Günther, Jackmin Ong, Isabelle Mohr, Alaeddine Abdessalem, Tanguy Abel, Mohammad~Kalim Akram, Susana Guzman, Georgios Mastrapas, Saba Sturua, Bo~Wang, Maximilian Werk, Nan Wang, and Han Xiao. 2024.
\newblock \href {http://arxiv.org/abs/2310.19923} {{Jina Embeddings 2: 8192-Token General-Purpose Text Embeddings for Long Documents}}.
\newblock \emph{arXiv:2310.19923}.

\bibitem[{Kurihara et~al.(2022)Kurihara, Kawahara, and Shibata}]{JGLUE}
Kentaro Kurihara, Daisuke Kawahara, and Tomohide Shibata. 2022.
\newblock \href {https://aclanthology.org/2022.lrec-1.317} {{JGLUE: Japanese General Language Understanding Evaluation}}.
\newblock In \emph{Proceedings of the Thirteenth Language Resources and Evaluation Conference (LREC)}, pages 2957--2966.

\bibitem[{Lee et~al.(2024{\natexlab{a}})Lee, Roy, Xu, Raiman, Shoeybi, Catanzaro, and Ping}]{NV-Embed}
Chankyu Lee, Rajarshi Roy, Mengyao Xu, Jonathan Raiman, Mohammad Shoeybi, Bryan Catanzaro, and Wei Ping. 2024{\natexlab{a}}.
\newblock \href {http://arxiv.org/abs/2405.17428} {{NV-Embed: Improved Techniques for Training LLMs as Generalist Embedding Models}}.
\newblock \emph{arXiv:2405.17428}.

\bibitem[{Lee et~al.(2024{\natexlab{b}})Lee, Dai, Ren, Chen, Cer, Cole, Hui, Boratko, Kapadia, Ding, Luan, Duddu, Abrego, Shi, Gupta, Kusupati, Jain, Jonnalagadda, Chang, and Naim}]{Gecko}
Jinhyuk Lee, Zhuyun Dai, Xiaoqi Ren, Blair Chen, Daniel Cer, Jeremy~R. Cole, Kai Hui, Michael Boratko, Rajvi Kapadia, Wen Ding, Yi~Luan, Sai Meher~Karthik Duddu, Gustavo~Hernandez Abrego, Weiqiang Shi, Nithi Gupta, Aditya Kusupati, Prateek Jain, Siddhartha~Reddy Jonnalagadda, Ming-Wei Chang, and Iftekhar Naim. 2024{\natexlab{b}}.
\newblock \href {http://arxiv.org/abs/2403.20327} {{Gecko: Versatile Text Embeddings Distilled from Large Language Models}}.
\newblock \emph{arXiv:2403.20327}.

\bibitem[{Li et~al.(2024{\natexlab{a}})Li, Ohagi, and Ri}]{JMTEB}
Shengzhe Li, Masaya Ohagi, and Ryokan Ri. 2024{\natexlab{a}}.
\newblock {J}{M}{T}{E}{B}: {J}apanese {M}assive {T}ext {E}mbedding {B}enchmark.
\newblock \url{https://huggingface.co/datasets/sbintuitions/JMTEB%7D%7D}.
\newblock [Accessed 31-08-2024].

\bibitem[{Li et~al.(2023)Li, Zhang, Zhang, Long, Xie, and Zhang}]{GTE}
Zehan Li, Xin Zhang, Yanzhao Zhang, Dingkun Long, Pengjun Xie, and Meishan Zhang. 2023.
\newblock \href {http://arxiv.org/abs/2308.03281} {{Towards General Text Embeddings with Multi-stage Contrastive Learning}}.
\newblock \emph{arXiv:2308.03281}.

\bibitem[{Li et~al.(2024{\natexlab{b}})Li, Zhang, Zhang, Long, Xie, and Zhang}]{mE5}
Zehan Li, Xin Zhang, Yanzhao Zhang, Dingkun Long, Pengjun Xie, and Meishan Zhang. 2024{\natexlab{b}}.
\newblock \href {http://arxiv.org/abs/2402.05672} {{Multilingual E5 Text Embeddings: A Technical Report}}.
\newblock \emph{arXiv:2402.05672}.

\bibitem[{Liu et~al.(2023)Liu, Xiao, Shao, and Cao}]{RetroMAE2}
Zheng Liu, Shitao Xiao, Yingxia Shao, and Zhao Cao. 2023.
\newblock \href {https://doi.org/10.18653/v1/2023.acl-long.148} {{RetroMAE-2: Duplex Masked Auto-Encoder For Pre-Training Retrieval-Oriented Language Models}}.
\newblock In \emph{Proceedings of the 61st Annual Meeting of the Association for Computational Linguistics (ACL)}, pages 2635--2648.

\bibitem[{Longpre et~al.(2021)Longpre, Lu, and Daiber}]{MKQA}
Shayne Longpre, Yi~Lu, and Joachim Daiber. 2021.
\newblock \href {https://doi.org/10.1162/tacl_a_00433} {{MKQA: A Linguistically Diverse Benchmark for Multilingual Open Domain Question Answering}}.
\newblock \emph{Transactions of the Association for Computational Linguistics (TACL)}, pages 1389--1406.

\bibitem[{Muennighoff et~al.(2023)Muennighoff, Tazi, Magne, and Reimers}]{MTEB}
Niklas Muennighoff, Nouamane Tazi, Loic Magne, and Nils Reimers. 2023.
\newblock \href {https://doi.org/10.18653/v1/2023.eacl-main.148} {{MTEB: Massive Text Embedding Benchmark}}.
\newblock In \emph{Proceedings of the 17th Conference of the European Chapter of the Association for Computational Linguistics (EACL)}, pages 2014--2037.

\bibitem[{Nussbaum et~al.(2024)Nussbaum, Morris, Duderstadt, and Mulyar}]{NOMIC}
Zach Nussbaum, John~X. Morris, Brandon Duderstadt, and Andriy Mulyar. 2024.
\newblock \href {http://arxiv.org/abs/2402.01613} {{Nomic Embed: Training a Reproducible Long Context Text Embedder}}.
\newblock \emph{arXiv:2402.01613}.

\bibitem[{Nvidia et~al.(2024)Nvidia, Adler, Agarwal, Aithal, Anh, Bhattacharya, Brundyn, Casper, Catanzaro, Clay, Cohen, Das, Dattagupta, Delalleau, Derczynski, Dong, Egert, Evans, Ficek, Fridman, Ghosh, Ginsburg, Gitman, Grzegorzek, Hero, Huang, Jawa, Jennings, Jhunjhunwala, Kamalu, Khan, Kuchaiev, LeGresley, Li, Liu, Liu, Long, Mahabaleshwarkar, Majumdar, Maki, Martinez, de~Melo, Moshkov, Narayanan, Narenthiran, Navarro, Nguyen, Nitski, Noroozi, Nutheti, Parisien, Parmar, Patwary, Pawelec, Ping, Prabhumoye, Roy, Saar, Sabavat, Satheesh, Scowcroft, Sewall, Shamis, Shen, Shoeybi, Sizer, Smelyanskiy, Soares, Sreedhar, Su, Subramanian, Sun, Toshniwal, Wang, Wang, You, Zeng, Zhang, Zhang, Zhang, Zhang, and Zhu}]{Nemotron}
Nvidia, Bo~Adler, Niket Agarwal, Ashwath Aithal, Dong~H. Anh, Pallab Bhattacharya, Annika Brundyn, Jared Casper, Bryan Catanzaro, Sharon Clay, Jonathan Cohen, Sirshak Das, Ayush Dattagupta, Olivier Delalleau, Leon Derczynski, Yi~Dong, Daniel Egert, Ellie Evans, Aleksander Ficek, Denys Fridman, Shaona Ghosh, Boris Ginsburg, Igor Gitman, Tomasz Grzegorzek, Robert Hero, Jining Huang, Vibhu Jawa, Joseph Jennings, Aastha Jhunjhunwala, John Kamalu, Sadaf Khan, Oleksii Kuchaiev, Patrick LeGresley, Hui Li, Jiwei Liu, Zihan Liu, Eileen Long, Ameya~Sunil Mahabaleshwarkar, Somshubra Majumdar, James Maki, Miguel Martinez, Maer~Rodrigues de~Melo, Ivan Moshkov, Deepak Narayanan, Sean Narenthiran, Jesus Navarro, Phong Nguyen, Osvald Nitski, Vahid Noroozi, Guruprasad Nutheti, Christopher Parisien, Jupinder Parmar, Mostofa Patwary, Krzysztof Pawelec, Wei Ping, Shrimai Prabhumoye, Rajarshi Roy, Trisha Saar, Vasanth Rao~Naik Sabavat, Sanjeev Satheesh, Jane~Polak Scowcroft, Jason Sewall, Pavel Shamis, Gerald Shen, Mohammad
  Shoeybi, Dave Sizer, Misha Smelyanskiy, Felipe Soares, Makesh~Narsimhan Sreedhar, Dan Su, Sandeep Subramanian, Shengyang Sun, Shubham Toshniwal, Hao Wang, Zhilin Wang, Jiaxuan You, Jiaqi Zeng, Jimmy Zhang, Jing Zhang, Vivienne Zhang, Yian Zhang, and Chen Zhu. 2024.
\newblock \href {http://arxiv.org/abs/2406.11704} {{Nemotron-4 340B Technical Report}}.
\newblock \emph{arXiv:2406.11704}.

\bibitem[{Okazaki et~al.(2024)Okazaki, Hattori, Shota, Iida, Ohi, Fujii, an~Mengsay~Loem, Yokota, and Mizuki}]{Swallow-Corpus}
Naoaki Okazaki, Kakeru Hattori, Hirai Shota, Hiroki Iida, Masanari Ohi, Kazuki Fujii, Taishi~Nakamura an~Mengsay~Loem, Rio Yokota, and Sakae Mizuki. 2024.
\newblock \href {https://openreview.net/forum?id=N5EYQSwW26} {{Building a Large Japanese Web Corpus for Large Language Models}}.
\newblock In \emph{Proceedings of the First Conference on Language Modeling (COLM)}, COLM.

\bibitem[{Press et~al.(2022)Press, Smith, and Lewis}]{ALiBi}
Ofir Press, Noah~A. Smith, and Mike Lewis. 2022.
\newblock \href {http://arxiv.org/abs/2108.12409} {{Train Short, Test Long: Attention with Linear Biases Enables Input Length Extrapolation}}.
\newblock \emph{arXiv:2108.12409}.

\bibitem[{Reimers and Gurevych(2019)}]{SentenceBERT}
Nils Reimers and Iryna Gurevych. 2019.
\newblock \href {https://arxiv.org/abs/1908.10084} {{Sentence-BERT: Sentence Embeddings using Siamese BERT-Networks}}.
\newblock In \emph{Proceedings of the 2019 Conference on Empirical Methods in Natural Language Processing and the 9th International Joint Conference on Natural Language Processing (EMNLP-IJCNLP)}, pages 3982--3992.

\bibitem[{Sato et~al.(2024)Sato, Tsukagoshi, Sasano, and Takeda}]{SomaSato}
Soma Sato, Hayato Tsukagoshi, Ryohei Sasano, and Koichi Takeda. 2024.
\newblock \href {https://aclanthology.org/2024.acl-srw.43} {{Improving Sentence Embeddings with Automatic Generation of Training Data Using Few-shot Examples}}.
\newblock In \emph{Proceedings of the 62nd Annual Meeting of the Association for Computational Linguistics Volume 4: Student Research Workshop (ACL SRW)}, pages 519--530.

\bibitem[{Shazeer(2020)}]{SwiGLU}
Noam Shazeer. 2020.
\newblock \href {http://arxiv.org/abs/2002.05202} {{GLU Variants Improve Transformer}}.
\newblock \emph{arXiv:2002.05202}.

\bibitem[{So et~al.(2022)So, Byun, Kang, and Cho}]{JaQuAD}
ByungHoon So, Kyuhong Byun, Kyungwon Kang, and Seongjin Cho. 2022.
\newblock \href {http://arxiv.org/abs/2202.01764} {{JaQuAD: Japanese Question Answering Dataset for Machine Reading Comprehension}}.
\newblock \emph{arXiv:2202.01764}.

\bibitem[{Su et~al.(2023)Su, Shi, Kasai, Wang, Hu, Ostendorf, Yih, Smith, Zettlemoyer, and Yu}]{InstructOR}
Hongjin Su, Weijia Shi, Jungo Kasai, Yizhong Wang, Yushi Hu, Mari Ostendorf, Wen-tau Yih, Noah~A. Smith, Luke Zettlemoyer, and Tao Yu. 2023.
\newblock \href {https://doi.org/10.18653/v1/2023.findings-acl.71} {{One Embedder, Any Task: Instruction-Finetuned Text Embeddings}}.
\newblock In \emph{Findings of the Association for Computational Linguistics: ACL 2023}, pages 1102--1121.

\bibitem[{Su et~al.(2024)Su, Ahmed, Lu, Pan, Bo, and Liu}]{RoPE}
Jianlin Su, Murtadha Ahmed, Yu~Lu, Shengfeng Pan, Wen Bo, and Yunfeng Liu. 2024.
\newblock \href {https://doi.org/https://doi.org/10.1016/j.neucom.2023.127063} {{RoFormer: Enhanced transformer with Rotary Position Embedding}}.
\newblock \emph{Neurocomputing}, 568:127063.

\bibitem[{Tateno(2024{\natexlab{a}})}]{JaCWIR}
Yuichi Tateno. 2024{\natexlab{a}}.
\newblock \href {https://huggingface.co/datasets/hotchpotch/JaCWIR} {{JaCWIR: Japanese Casual Web IR - 日本語情報検索評価のための小規模でカジュアルなWebタイトルと概要のデータセット}}.

\bibitem[{Tateno(2024{\natexlab{b}})}]{JQaRA}
Yuichi Tateno. 2024{\natexlab{b}}.
\newblock \href {https://huggingface.co/datasets/hotchpotch/JQaRA} {{JQaRA: Japanese Question Answering with Retrieval Augmentation - 検索拡張(RAG)評価のための日本語Q\&Aデータセット}}.

\bibitem[{Tsukagoshi et~al.(2023)Tsukagoshi, Sasano, and Takeda}]{JapaneseSimCSETechReport}
Hayato Tsukagoshi, Ryohei Sasano, and Koichi Takeda. 2023.
\newblock \href {http://arxiv.org/abs/2310.19349} {{Japanese SimCSE Technical Report}}.
\newblock \emph{arXiv:2310.19349}.

\bibitem[{Wang et~al.(2022)Wang, Yang, Huang, Jiao, Yang, Jiang, Majumder, and Wei}]{E5}
Liang Wang, Nan Yang, Xiaolong Huang, Binxing Jiao, Linjun Yang, Daxin Jiang, Rangan Majumder, and Furu Wei. 2022.
\newblock \href {http://arxiv.org/abs/2212.03533} {{Text Embeddings by Weakly-Supervised Contrastive Pre-training}}.
\newblock \emph{arXiv:2212.03533}.

\bibitem[{Wang et~al.(2023)Wang, Yang, Huang, Jiao, Yang, Jiang, Majumder, and Wei}]{SimLM}
Liang Wang, Nan Yang, Xiaolong Huang, Binxing Jiao, Linjun Yang, Daxin Jiang, Rangan Majumder, and Furu Wei. 2023.
\newblock \href {https://doi.org/10.18653/v1/2023.acl-long.125} {{SimLM: Pre-training with Representation Bottleneck for Dense Passage Retrieval}}.
\newblock In \emph{Proceedings of the 61st Annual Meeting of the Association for Computational Linguistics (ACL)}, pages 2244--2258.

\bibitem[{Wang et~al.(2024)Wang, Yang, Huang, Yang, Majumder, and Wei}]{MistralE5}
Liang Wang, Nan Yang, Xiaolong Huang, Linjun Yang, Rangan Majumder, and Furu Wei. 2024.
\newblock \href {https://aclanthology.org/2024.acl-long.642} {{Improving Text Embeddings with Large Language Models}}.
\newblock In \emph{Proceedings of the 62nd Annual Meeting of the Association for Computational Linguistics (ACL)}, pages 11897--11916.

\bibitem[{Williams et~al.(2018)Williams, Nangia, and Bowman}]{MNLI}
Adina Williams, Nikita Nangia, and Samuel Bowman. 2018.
\newblock \href {https://doi.org/10.18653/v1/N18-1101} {{A Broad-Coverage Challenge Corpus for Sentence Understanding through Inference}}.
\newblock In \emph{Proceedings of the 2018 Conference of the North American Chapter of the Association for Computational Linguistics: Human Language Technologies (NAACL)}, pages 1112--1122.

\bibitem[{Xiao et~al.(2022)Xiao, Liu, Shao, and Cao}]{RetroMAE}
Shitao Xiao, Zheng Liu, Yingxia Shao, and Zhao Cao. 2022.
\newblock \href {https://doi.org/10.18653/v1/2022.emnlp-main.35} {{RetroMAE: Pre-Training Retrieval-oriented Language Models Via Masked Auto-Encoder}}.
\newblock In \emph{Proceedings of the 2022 Conference on Empirical Methods in Natural Language Processing (EMNLP)}, pages 538--548.

\bibitem[{Xiao et~al.(2024)Xiao, Liu, Zhang, and Muennighoff}]{BGE}
Shitao Xiao, Zheng Liu, Peitian Zhang, and Niklas Muennighoff. 2024.
\newblock \href {https://arxiv.org/abs/2309.07597} {{C-Pack: Packaged Resources To Advance General Chinese Embedding}}.
\newblock In \emph{The 47th International ACM SIGIR Conference on Research and Development in Information Retrieval (SIGIR)}.

\bibitem[{Yanaka and Mineshima(2021)}]{JaNLI}
Hitomi Yanaka and Koji Mineshima. 2021.
\newblock \href {https://doi.org/10.18653/v1/2021.blackboxnlp-1.26} {{Assessing the Generalization Capacity of Pre-trained Language Models through {J}apanese Adversarial Natural Language Inference}}.
\newblock In \emph{Proceedings of the Fourth BlackboxNLP Workshop on Analyzing and Interpreting Neural Networks for NLP (BlackboxNLP)}, pages 337--349.

\bibitem[{Zhang et~al.(2021{\natexlab{a}})Zhang, Li, Xiao, Zhu, Nallapati, Arnold, and Xiang}]{PairSupCon}
Dejiao Zhang, Shang-Wen Li, Wei Xiao, Henghui Zhu, Ramesh Nallapati, Andrew~O. Arnold, and Bing Xiang. 2021{\natexlab{a}}.
\newblock \href {https://doi.org/10.18653/v1/2021.emnlp-main.467} {{Pairwise Supervised Contrastive Learning of Sentence Representations}}.
\newblock In \emph{Proceedings of the 2021 Conference on Empirical Methods in Natural Language Processing (EMNLP)}, pages 5786--5798.

\bibitem[{Zhang et~al.(2023)Zhang, Lan, and He}]{SynCSE}
Junlei Zhang, Zhenzhong Lan, and Junxian He. 2023.
\newblock \href {https://doi.org/10.18653/v1/2023.emnlp-main.238} {{Contrastive Learning of Sentence Embeddings from Scratch}}.
\newblock In \emph{Proceedings of the 2023 Conference on Empirical Methods in Natural Language Processing (EMNLP 2023)}, pages 3916--3932.

\bibitem[{Zhang et~al.(2021{\natexlab{b}})Zhang, Ma, Shi, and Lin}]{MrTyDi}
Xinyu Zhang, Xueguang Ma, Peng Shi, and Jimmy Lin. 2021{\natexlab{b}}.
\newblock \href {https://doi.org/10.18653/v1/2021.mrl-1.12} {{Mr. TyDi: A Multi-lingual Benchmark for Dense Retrieval}}.
\newblock In \emph{Proceedings of the 1st Workshop on Multilingual Representation Learning (MRL)}, pages 127--137.

\bibitem[{Zhang et~al.(2022)Zhang, Thakur, Ogundepo, Kamalloo, Alfonso-Hermelo, Li, Liu, Rezagholizadeh, and Lin}]{MIRACL}
Xinyu Zhang, Nandan Thakur, Odunayo Ogundepo, Ehsan Kamalloo, David Alfonso-Hermelo, Xiaoguang Li, Qun Liu, Mehdi Rezagholizadeh, and Jimmy Lin. 2022.
\newblock \href {http://arxiv.org/abs/2210.09984} {{Making a MIRACL: Multilingual Information Retrieval Across a Continuum of Languages}}.
\newblock \emph{arXiv:2210.09984}.

\bibitem[{吉越 et~al.(2020)吉越, 河原, and 黒橋}]{JSNLI}
卓見 吉越, 大輔 河原, and 禎夫 黒橋. 2020.
\newblock \href {https://ipsj.ixsq.nii.ac.jp/ej/index.php?active_action=repository_view_main_item_detail&page_id=13&block_id=8&item_id=206114&item_no=1} {{機械翻訳を用いた自然言語推論データセットの多言語化}}.
\newblock \emph{第２４４回自然言語処理研究会（ＮＬ研）}.

\bibitem[{鈴木 et~al.(2020)鈴木, 鈴木, 松田, ⻄田, and 井之上}]{JAQKET}
正敏 鈴木, 潤 鈴木, 耕史 松田, 京介 ⻄田, and 直也 井之上. 2020.
\newblock \href {https://www.anlp.jp/proceedings/annual_meeting/2020/pdf_dir/P2-24.pdf} {{JAQKET: クイズを題材にした日本語QAデータセットの構築}}.
\newblock In \emph{言語処理学会第26回年次大会}.

\end{thebibliography}

\appendix

\section{Hyperparameters}
\label{appendix:hyperparameters}
Table~\ref{tab:hyperparameters-emb} shows the hyperparameters and settings used during the training of the embedding model, and Table~\ref{tab:hyperparameters-reranker} shows the hyperparameters and settings used during the training of the reranker.
\begin{table*}[t]
\small
\centering
\begin{tabular}{l@{\hspace{6ex}}ccc@{\hspace{8ex}}ccc}
\bhline
\tabH Phase & \multicolumn{3}{c}{Pre-training} & \multicolumn{3}{c}{Fine-tuning} \\
Model & \RuriPTSmall & \RuriPTBase  & \RuriPTLarge & \RuriSmall & \RuriBase & \RuriLarge \\
\bhline
\tabH learning rate & 1$\times10^{-4}$ & 5$\times10^{-5}$ & 3$\times10^{-5}$ & 1$\times10^{-5}$ & 5$\times10^{-6}$ & 3$\times10^{-6}$  \\
max length & 256 & 256 & 192 & 512 & 512 & 512 \\
warmup ratio & 10\% & 10\% & 10\% & 10\% & 10\% & 10\% \\
batch size & 8192 & 8192 & 8192 & 512 & 512& 512 \\
epochs & 1 & 1 & 1 & 1 & 1 & 1 \\
$\tau$ & 0.01 & 0.01 & 0.01 & 0.01 & 0.01 & 0.01 \\
weight decay & 0.01 & 0.01 & 0.01 & 0.01 & 0.01 & 0.01 \\
hard negatives & 1 & 1 & 1 & 15 & 15 & 15 \\
task-homogeneous & \checkmark & \checkmark & \checkmark & \checkmark & \checkmark & \checkmark \\
shuffle positive & \checkmark & \checkmark & \checkmark & \checkmark & \checkmark & \checkmark \\
knowledge distillation &  & &  & \checkmark & \checkmark & \checkmark \\
\bhline
\end{tabular}
\caption{
Hyperparameters for contrastive pre-training and supervised fine-tuning.
}
\label{tab:hyperparameters-emb}
\end{table*}
\begin{table*}[t]
\small
\centering
\begin{tabular}{l@{\hspace{6ex}}ccc@{\hspace{8ex}}ccc}
\bhline
\tabH Phase & \multicolumn{3}{c}{Stage1} & \multicolumn{3}{c}{Stage2} \\
Model & Small & Base & Large & Small & Base & Large \\
\bhline
\tabH learning rate & 1$\times10^{-4}$ & 5$\times10^{-5}$ & 3$\times10^{-5}$ & 1$\times10^{-5}$ & 5$\times10^{-6}$ & 3$\times10^{-6}$  \\
max length & 256 & 256 & 256 & 512 & 512 & 512 \\
warmup ratio & 10\% & 10\% & 10\% & 10\% & 10\% & 10\% \\
batch size & 512 & 512 & 512 & 64 & 64 & 64\\
epochs & 1 & 1 & 1 & 1 & 1 & 1 \\
weight decay & 0.01 & 0.01 & 0.01 & 0.01 & 0.01 & 0.01 \\
hard negatives & 63 & 63 & 63 & 63 & 63 & 63 \\
task-homogeneous  &  &  &  &  &  &  \\
shuffle positive & \checkmark & \checkmark & \checkmark & & & \\
\bhline
\end{tabular}
\caption{
Hyperparameters for rerankers.
}
\label{tab:hyperparameters-reranker}
\end{table*}

\end{document}